\documentclass[10pt,twocolumn,letterpaper]{article}

\usepackage{cvpr}              

\usepackage{amssymb}
\usepackage{amsthm}
\usepackage{algorithm}
\usepackage{algorithmic}

\usepackage{booktabs}       
\usepackage{amsfonts}       
\usepackage{nicefrac}       
\usepackage{microtype}      
\usepackage{xcolor}         
\usepackage{amsmath}
\usepackage{amsthm}
\usepackage{multirow}
\usepackage{colortbl}
\definecolor{lightpastelpurple}{rgb}{0.69, 0.61, 0.85}
\colorlet{LightGreen}{lightpastelpurple!40}
\usepackage{tabularx}
\usepackage{tikz}

\definecolor{cvprblue}{rgb}{0.21,0.49,0.74}
\usepackage[breaklinks,colorlinks,allcolors=cvprblue]{hyperref}

\def\paperID{38416} 
\def\confName{CVPR}
\def\confYear{2026}

\title{\textsc{SouPLe}: Enhancing Audio-Visual Localization and Segmentation with Learnable Prompt Contexts}

\author{Khanh Binh Nguyen\\
Deakin University, Australia\\
{\tt\small binh.nguyen@deakin.edu.au}
\and
Chae Jung Park\thanks{Corresponding author.}\\
National Cancer Center, South Korea\\
{\tt\small cjp@ncc.re.kr}
}

\begin{document}
\maketitle
\begin{abstract}
Large-scale pre-trained image-text models exhibit robust multimodal representations, yet applying the Contrastive Language-Image Pre-training (CLIP) model to audio-visual localization remains challenging. Replacing the classification token ($[CLS]$) with an audio-embedded token ($[V_A]$) struggles to capture semantic cues, and the prompt “a photo of a $[V_A]$” fails to establish meaningful connections between audio embeddings and context tokens. To address these issues, we propose Sound-aware Prompt Learning (\textsc{SouPLe}), which replaces fixed prompts with learnable context tokens. These tokens incorporate visual features to generate conditional context for a mask decoder, effectively bridging semantic correspondence between audio and visual inputs. Experiments on VGGSound, SoundNet, and AVSBench demonstrate that \textsc{SouPLe} improves localization and segmentation performance.
\end{abstract}    

\section{Introduction}
\label{sec:intro}
Localizing sound sources within visual scenes is a key component of audiovisual perception, which is vital for both biological organisms and artificial systems.
The domain of audio-visual sound-source localization has experienced considerable progress in recent years, propelled by the imperative to create machine perception systems capable of emulating human multi-sensory integration for pinpointing the origins of sound in intricate settings.
In recent years, audio-visual sound source localization has been extensively explored to equip machine perception with comparable capabilities \cite{arandjelovic2018objects,chen2021localizing,hu2020discriminative,li2021space,lin2023unsupervised,liu2022exploiting,mo2022closer,mo2022localizing,park2023marginnce,qian2020multiple,senocak2018learning,senocak2022learning,senocak2022less,senocak2023sound,song2022self,sun2023learning,nguyen2024save}.
A key strategy is to utilize the inherent correlation between audio and visual signals without explicit supervision or annotated data.
The primary method for accomplishing this aligns audio-visual representations as self-supervision signals within a contrastive learning framework.

\begin{figure}[!t]
    \centering
    \includegraphics[width=0.85\linewidth]{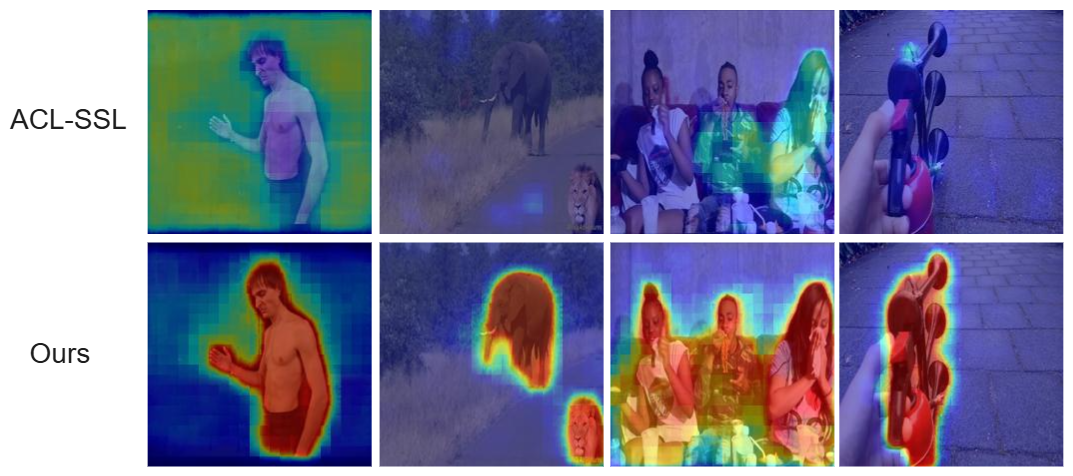}
    \caption{Comparison of attention masks produced by ACL-SSL and \textsc{SouPLe} in representative cases. Compared with ACL-SSL, \textsc{SouPLe} yields more refined localization of the sounding objects.}
    \label{fig:acl-failure}
\end{figure}

Sound source localization methods often incorporate additional prior knowledge beyond the fundamental assumption of audio-visual correspondence.
This prior knowledge can take various forms, such as visual object detection \cite{mo2022closer,mo2022localizing} or motion analysis \cite{xuan2022proposal}.
However, these priors may introduce biases that can potentially hinder true audio-visual semantic alignment, which is crucial for accurate sound source localization \cite{arandjelovic2018objects,mo2022closer,oya2020we}.
\citet{Park_2024_WACV} proposed a CLIP-based adaptation method that leverages the audio-driven embedding to efficiently localize and segment sounding objects.
Nevertheless, ACL-SSL \cite{Park_2024_WACV} fails in some cases, as depicted in Figure \ref{fig:acl-failure}.
We conjecture that this is because only the classification token ($[CLS]$) is replaced with an audio-embedded token ($[V_A]$) that does not have semantic information that can be integrated with visual information. 
Moreover, the prompt ``a photo of a $[V_A]$" does not always hold, and the given tokens of the phrase ``a photo of a" are not well aligned with $[V_A]$.

Our research adopted a different approach by leveraging the prompt learning approach to enhance the generalization of CLIP-based methods such as ACL-SSL.
Prompt learning was proposed to overcome the limitations of prompt engineering \cite{zhou2022learning,zhou2022conditional,khattakMaPLe,Khattak_2023_ICCV, 10870225}. 
In prompt learning, the prompt tokens are trained using few-shot labeled data for downstream tasks, freezing the vision-language model (VLM) image and text encoders.
Context optimization (CoOp) \cite{zhou2022learning}, a milestone in prompt learning, defines prompts as learnable vectors and optimizes them to fit few-shot training samples effectively.
In addition, to address the limited generalization of CoOp to downstream tasks, conditional context optimization (CoCoOp) \cite{zhou2022conditional} was introduced as a successor model.
CoCoOp generates prompts conditioned on image features, thereby yielding improved generalization capabilities for unseen categories.

Building on these observations, we propose \textbf{Sou}nd-aware \textbf{P}rompt \textbf{Le}arning (\textsc{SouPLe}), a prompt learning framework for audio-visual localization and segmentation. Instead of relying on a fixed handcrafted prompt, \textsc{SouPLe} introduces learnable context tokens that adapt to the input visual features and provide instance-conditional context for the audio representation. This design strengthens the semantic correspondence between audio and visual modalities, leading to improved localization and segmentation performance. Our contributions are summarized as follows:
 \begin{itemize}
    \item We propose a prompt learning framework to address the generalization limitations of CLIP-based audio-visual localization by introducing instance-conditional, learnable context tokens.
    \item We present an end-to-end label-free framework that generalizes to unseen objects and datasets without requiring ground-truth class labels.
    \item Our prompt learning strategy produces context-aware tokens that are combined with an audio-embedded token to better emphasize sound-associated regions in the visual input.
    \item Extensive experiments on VGG-SS, SoundNet-Flickr, VGG-SS OpenSet, AVS-Bench, and Extended VGG-SS/SoundNet-Flickr demonstrate the effectiveness of the proposed method.
 \end{itemize}
\section{Related Work}
\label{sec:rw}
\paragraph{Sound Source Localization}
The leading approach for audio-visual sound source localization is cross-modal attention \cite{senocak2018learning,senocak2022learning,tian2018audio}, which is typically combined with contrastive loss.
In line with the contrastive learning approach, advancements include the integration of hard negatives from background areas \cite{chen2021localizing}, use of iterative contrastive learning with pseudo-labels from prior epochs \cite{lin2023unsupervised}, enforcement of transformation invariance and equivariance via data augmentation and geometric consistency \cite{liu2022exploiting}, selection of semantically similar hard positives \cite{senocak2022learning}, adoption of negative-free contrastive learning \cite{song2022self} akin to SiamSiam \cite{chen2021exploring}, employing momentum encoders to prevent overfitting \cite{mo2022closer}, introduction of a negative margin to contrastive learning to reduce the impact of noisy correspondences \cite{park2023marginnce}, and implementation of false negative-aware contrastive learning through intra-modal similarities \cite{sun2023learning}.

While GAVS \cite{wang2023prompting} employs prompt learning to directly condition the segmentation decoder, \textsc{SouPLe} differs in both where and how prompting is applied. Specifically, \textsc{SouPLe} performs soft prompt learning within a CLIP-based, textless grounding pipeline by replacing the static handcrafted prompt (e.g., "a photo of a $[V_A]$") with instance-conditional context tokens generated from visual features via a Meta-net. We therefore view the comparison to GAVS as a cross-paradigm reference, while emphasizing that our main contribution is prompt learning within a CLIP-based audio-visual grounding framework.

In addition, various sound localization methods that utilize extra prior knowledge or post-processing techniques are being explored.
Label information has been integrated to train core audio and visual networks, improving alignment between audio and visual cues \cite{qian2020multiple,senocak2022less}. Object priors in the form of object proposals have also been applied, while post-processing methods using pre-trained visual feature activation maps can enhance audio-visual localization results \cite{xuan2022proposal}.
In our research, we employ the multimodal alignment capabilities of CLIP as a prior in a textless, entirely self-supervised approach, foregoing any post-processing steps.

\paragraph{CLIP in Audio-Visual Learning}
Recent advancements in CLIP models pre-trained on extensive paired datasets \cite{jia2021scaling,radford2021learning}, have shown remarkable generalization capabilities, proving effective in a variety of downstream tasks across diverse research areas.
This section discusses studies that integrate CLIP \cite{radford2021learning} into audiovisual learning, using WAV2CLIP \cite{wu2022wav2clip} and AudioCLIP \cite{guzhov2022audioclip} to extend the pretrained CLIP model by synchronizing audio features with text and visual features within a unified embedding space, thereby facilitating representation learning.
Synchronization is achieved through paired data or by leveraging the visual modality as an intermediary.
Beyond representation learning, CLIP models are utilized in tasks such as audio-visual event localization \cite{mahmud2023ave}, video parsing \cite{fan2024revisit}, and audio-visual source separation \cite{dong2022clipsep,tan2023language}.
Notably, while \cite{tan2023language} used text input for separation, CLIPSep \cite{dong2022clipsep} was trained to focus on audiovisual correlation, omitting text queries.
The method proposed in this study similarly focuses on training with an audio-visual alignment goal.
Additionally, other studies \cite{bhati2023segmental,yariv2023audiotoken} modified pretrained CLIP models and text encoders to process audio by replicating contextual text tokens with audio signals, thus enabling the CLIP text encoder to process audio signals.
Our research adopted a comparable strategy, utilizing the CLIP model without text input for sound localization tasks.

\paragraph{Prompt Learning}
Drawing inspiration from ``prompt engineering" in natural language processing (NLP), ``prompt learning" seeks to derive optimal prompts by leveraging few-shot samples in downstream tasks.
The core methodology of prompt learning involves optimizing prompts that are represented as continuous learnable vectors using few-shot samples.
Typically, the cross-entropy loss function serves as the training objective.
CoOp \cite{zhou2022learning} was a trailblazing effort in conceptualizing prompts as continuous vectors instead of discrete entities.
Building on this, CoCoOp \cite{zhou2022conditional} adapted prompts to image features to improve generalization across unseen categories.
MaPLe \cite{khattakMaPLe} extends prompt learning to multimodal settings by injecting learnable vectors into both the text and image encoders,
enabling more effective multimodal adaptation.
PromptSRC \cite{Khattak_2023_ICCV} addresses the crucial aspect of knowledge retention during pretraining by introducing a sophisticated prompt-learning model that balances both task-agnostic and task-specific knowledge.

\subsection{Prompt Learning for CLIP}
Prompt learning eliminates the need for the handcrafted design of prompts, e.g., "a photo of a", to match downstream tasks.
The earliest work, CoOp \cite{zhou2022learning}, deﬁnes a prompt as sequence of $M$ continuously differentiable tokens, $[V_1][V_2] \dots [V_M]$.
In the case of CLIP-ViT, $[V_i]$ is a 512-dimensional vector.
The prompt representing the i-th category can be then deﬁned as $t_i{\left(x\right)} = \{V_1{\left(x\right)}, V_2{\left(x\right)},\dots, V_M{\left(x\right)}, c_i\}$.
Let features from the image encoder and text encoder be denoted by $x$ and $g(\cdot)$.
Then the class probabilities can be expressed by the following formula:
\begin{equation}
    p(\hat{y} \mid \mathbf{x})=\frac{\exp \left(\operatorname{sim}\left(\mathbf{x}, g\left(\mathbf{t}_y\right)\right) / \tau\right)}{\sum_{i=1}^C \exp \left(\operatorname{sim}\left(\mathbf{x}, g\left(\mathbf{t}_i\right)\right) / \tau\right)}
\end{equation}
Here, $\text{sim}\left(\cdot, \cdot\right)$ is a metric that measures similarity in a feature space, with the cosine similarity as a common choice, and $\tau$ is the temperature parameter.
CoCoOp \cite{zhou2022conditional} conditions prompts with image features to enhance generalization for unseen categories.
In practical, image-conditioned prompts, $t_i{\left(x\right)} = \{V_1{\left(x\right)}, V_2{\left(x\right)},\dots , V_M{\left(x\right)}, c_i\}$ are formulated by summing meta-tokens $\pi$, derived from "meta-net" $\theta$, and the $[V_i]$.
The class probabilities are expressed by
\begin{equation}
    p(\hat{y} \mid \mathbf{x})=\frac{\exp \left(\operatorname{sim}\left(\mathbf{x}, g\left(\mathbf{t}_y(\mathbf{x})\right)\right) / \tau\right)}{\sum_{i=1}^C \exp \left(\operatorname{sim}\left(\mathbf{x}, g\left(\mathbf{t}_i(\mathbf{x})\right)\right) / \tau\right)} .
\end{equation}
Both CoOp and CoCoOp update tokens; CoCoOp additionally adjusts the meta-net parameters - by using the cross-entropy loss from downstream tasks:
\begin{equation}
    \mathcal{L}_{\text{ce}}(y,\hat{y})=-\sum_{i=1}^C y_i \text{log}\left(\hat{y}_i\right)
\end{equation}

\subsection{Sound Source Localization Using CLIP}
\citet{Park_2024_WACV} introduce a framework named ACL-SSL that translates audio signals into tokens compatible with CLIP's text encoder, yielding audio-driven embeddings.
The method generates audio-grounded masks for the provided audio using these audio-driven embeddings.
On the other hand, the model extracts audio-grounded image features from the highlighted regions in the visual input.
The extracted features are aligned with the audio-driven embeddings using an audio-visual correspondence objective.
Finally, the entire model is trained using a contrastive learning framework, which helps in learning the audio-visual correspondence without explicit text input.

ACL-SSL leverages the pre-trained CLIP model's robust representational capabilities and effective multimodal alignment but uniquely does so without using explicit text input.
Instead, it relies solely on audio-visual correspondence.
The use of pre-trained image-text models enables the generation of improved localization maps for sounding objects.

\begin{figure*}[!t]
    \includegraphics[width=\linewidth]{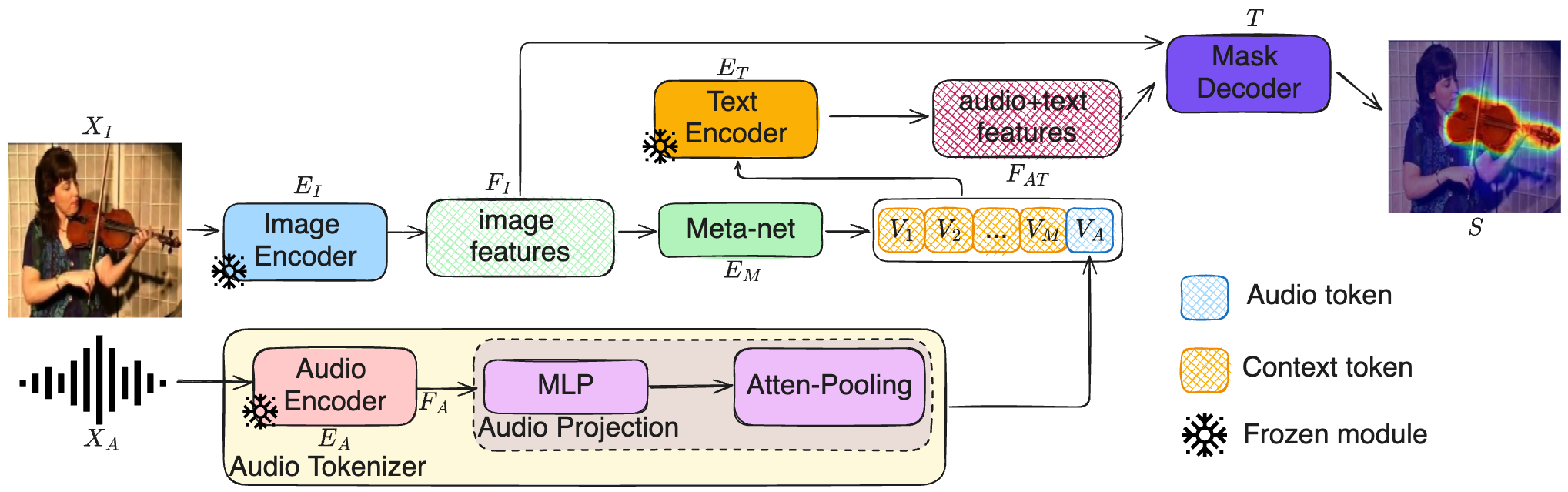}
    \caption{Pipeline of our \textsc{SouPLe} framework. \textsc{SouPLe} takes an audio-visual pair as input, converting the audio signal into a CLIP-compatible token via the Audio Tokenizer and the image into learnable context tokens via the Meta-net. The learnable context tokens are concatenated with the audio-embedded token and used by the text encoder to produce audio-text features for mask decoding. By replacing fixed handcrafted prompts with learnable context tokens, \textsc{SouPLe} improves generalization to the given audio-visual input.}
    \label{fig:pipeline}
\end{figure*}

\section{Method}
Our method builds upon ACL-SSL \cite{Park_2024_WACV} and extends it with prompt learning, as illustrated in Figure \ref{fig:pipeline}.

\subsection{SOUPLE Framework}
Using CLIPSeg \cite{luddecke2022image} as the audio-visual grounder, an audio-visual pair is input to the framework.
The CLIP Image Encoder $E_I$ extracts the image features $F_I$ from the visual input $X_I$.
Then, a meta-net $E_M$, which consists of a two-layer non-linear bottleneck structure (Linear-ReLU-Linear), with the hidden layer reducing the input dimension by $16\times$, receives these image features and translates them into $M$ context tokens ($[V_1][V_2] \dots [V_M]$).

On the other hand, the Audio Encoder $E_A$ extracts the audio features $F_A$ from the audio signal $X_A$ and translates them into the compatible audio-embedded token $[V_A]$ via the Audio Projection module, which consists of an MLP and an attention pooling layer.
This module is referred to as the Audio Tokenizer.
The context tokens and the audio-embedded token are then concatenated and input to the Text Encoder $E_T$ to extract the audio-text features $F_{AT}$.
Finally, image features and audio-text features are input into the Mask Decoder $D$, providing the segmentation masks $S$ for given sounding objects, following the mechanism of ACL-SSL \cite{Park_2024_WACV}.

In this way, instead of feeding the fixed tokens of a prompt "a photo of a $[V_A]$", we replace them with the learnable context tokens $[V_1][V_2] \dots [V_M][V_A]$.
This improves generalization for the given input by introducing instance-conditional context, since the fixed prompt “a photo of a” has no semantic connection to $[V_A]$.

\subsection{Visual-Audio-Text Alignment}
\begin{figure}[!t]
    \includegraphics[width=\linewidth]{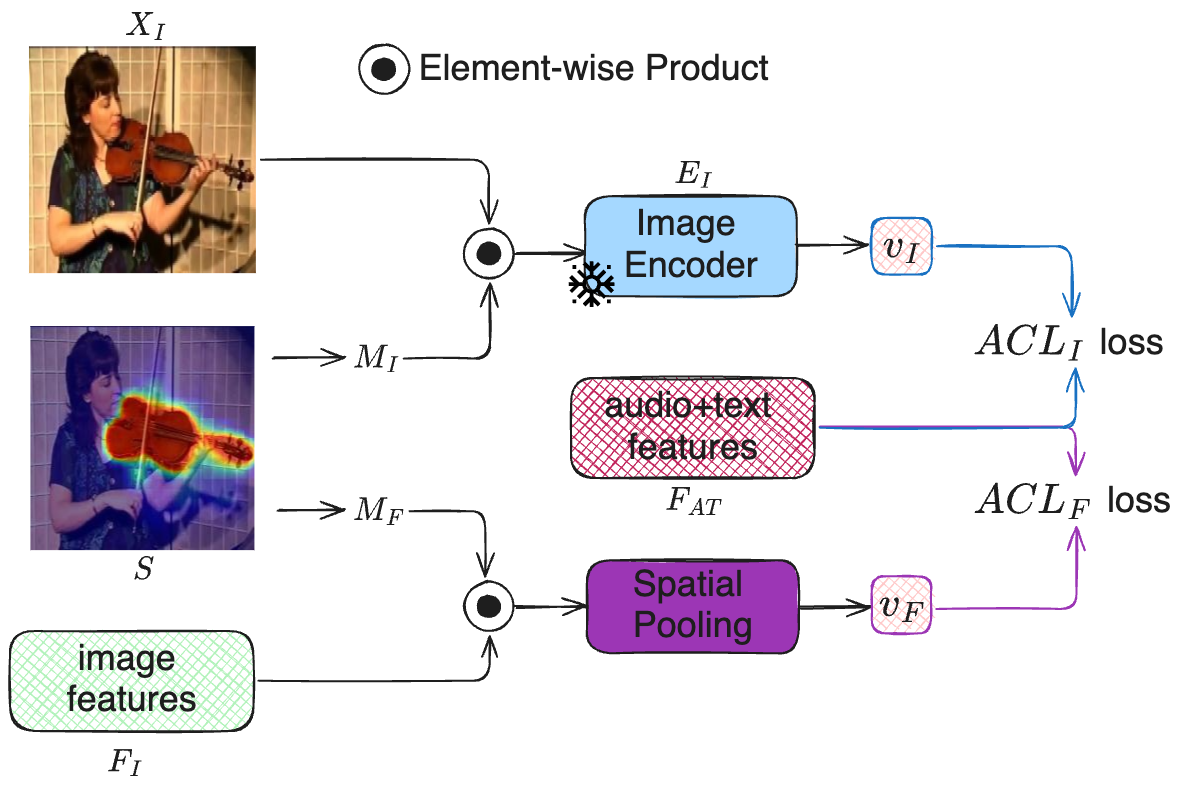}
    \caption{Visual-Audio-Text Alignment (VAT) module}
    \label{fig:decoder}
\end{figure}

Figure \ref{fig:decoder} illustrates the computation of contrastive loss at both the image and feature levels by the Visual-Audio-Text (VAT) module.
Sounding area masks derived from the audio input using \textsc{SouPLe} are utilized to create two distinct masks: one for the image level ($M_I$), which highlights the sounding regions by foregrounding pixels and obscuring the background, and another for the feature level ($M_F$), which accentuates areas within the spatial visual features.
The image mask is then converted into a visual embedding, denoted as $v_I = E_I(M_I \cdot X_I)$.
In a similar fashion, the visual embedding for spatial visual features is formulated as $v_I = M_I \cdot F_I$.
Subsequently, the VAT similarity between the audio-text features ($F_{AT}$) and the audio-grounded visual embedding $v_I$ is determined through cosine similarity, expressed as $S^I = (v_I \cdot F_{AT})$.
Likewise, the similarity between the audio-text features and the feature-level audio-grounded visual embedding $v_F$ is articulated as $S^F = (v_F \cdot F_{AT})$.
Symmetric InfoNCE is applied to calculate the image-level audio-text-grounded contrastive loss ($\mathcal{L}_{ACL_I}$) and the feature-level audio-text-grounded contrastive loss ($\mathcal{L}_{ACL_F}$).

\begin{equation}
    \begin{aligned}
    \mathcal{L}_{ACL_I} &= InfoNCE\left(S^I\right) \\
    = & -\frac{1}{2 B} \sum_i^B \log \frac{\exp \left(S_{i, i}^I / \tau\right)}{\sum_j^B \exp \left(S_{i, j}^I / \tau\right)} \\
    & -\frac{1}{2 B} \sum_i^B \log \frac{\exp \left(S_{i, i}^I / \tau\right)}{\sum_j^B \exp \left(S_{j, i}^I / \tau\right)}
    \end{aligned}
\end{equation}
\begin{equation}
    \begin{aligned}
    \mathcal{L}_{ACL_F} &= InfoNCE\left(S^F\right) \\
    = & -\frac{1}{2 B} \sum_i^B \log \frac{\exp \left(S_{i, i}^F / \tau\right)}{\sum_j^B \exp \left(S_{i, j}^F / \tau\right)} \\
    & -\frac{1}{2 B} \sum_i^B \log \frac{\exp \left(S_{i, i}^F / \tau\right)}{\sum_j^B \exp \left(S_{j, i}^F / \tau\right)}
    \end{aligned}
\end{equation}
where $\tau$ is the temperature parameter and $S_I$ is image-level
audio-visual similarity matrix within batch $B$.

Finally, following ACL-SSL, the area regularization loss ($\mathcal {L}_{REG}$) is defined as:
\begin{equation}
    \mathcal {L}_{REG}= \sum_i{\left\|p^+ - \hat{M^I_{i,i}}\right\|_1} + \sum_{i\neq j} \left\|p^- - \hat{M^I_{i,j}} \right\|_1
\end{equation}

\subsection{Training Objectives}
The overall training loss term is following ACL-SSL, as defined as follows:
\begin{equation}
    \label {eq:final} 
    \mathcal {L} = \lambda_1 \mathcal {L}_{ACL_I} + \lambda_2 \mathcal {L}_{ACL_F} + \lambda_3 \mathcal {L}_{REG}
\end{equation}
where $\lambda_1$, $\lambda_2$, and $\lambda_3$ are the hyper-parameters weighting the loss terms.
The $\mathcal {L}_{ACL_I}$ loss represents the symmetric InfoNCE form of contrastive loss.
It enhances the similarity between the positive-sounding region and its corresponding audio pair while simultaneously ensuring the dissimilarity between negative audio samples and the actual-sounding region.
The $\mathcal {L}_{ACL_F}$ loss focuses on the highly correlated area, regardless of positive or negative audio-visual pairs.
The $\mathcal {L}_{REG}$ loss regularizes the size of the mask during training to ensure it only covers the relevant sounding area.
Except for the encoders, $E_I$, $E_A$, and $E_T$ parameters are frozen; the rest of the framework is optimized using Equation \ref{eq:final} as the overall training objective.
\section{Experiments}
\paragraph{Datasets} Our method utilizes the VGGSound dataset \cite{chen2020vggsound}, which consists of approximately 200,000 videos. Post-training, we assess the sound localization capabilities on the VGG-SS \cite{chen2021localizing} and SoundNet-Flickr \cite{senocak2018learning,senocak2022learning} datasets. These datasets offer bounding box annotations for sound-emitting objects, with around 5,000 and 250 samples, respectively.
Additional evaluations are performed on the AVSBench \cite{zhou2022audio} and Extended VGG-SS/SoundNet-Flickr \cite{mo2022closer} datasets.
AVSBench provides binary segmentation maps that identify pixels associated with audio-visual signals and are segmented into Single-source (S4) and Multi-source (MS3) categories based on the number of sound sources.
The S4 category contains approximately 5,000 samples, while the MS3 has about 400 samples.
Finally, the Extended VGG-SS/SoundNet-Flickr datasets, introduced by \citet{mo2022closer}, are employed to investigate the presence of sound sources that are not visible.

\paragraph{Implementation Details} In our approach, we utilize a frozen pre-trained ViT-B/16 CLIP model as the image encoder, BEATs \cite{chen2022beats} for the audio encoder, and CLIPSeg \cite{lueddecke22_cvpr} as the grounder, following the design of ACL-SSL \cite{Park_2024_WACV}.
For training, 10-second audio clips sampled at 16kHz were employed, along with the central video frame resized to $352\times352$.
The model underwent optimization over 20 epochs with a batch size of 16, utilizing the Adam optimizer with a learning rate of $10^{-3}$ and a weight decay of $10^{-5}$.
The prompt-learning components add only a small parameter overhead (approximately 2.38M trainable parameters in total, less than 1\% of the overall model capacity of $\sim 242\,\text{M}$).

\subsection{Quantitative Results}
We compare our proposed method with the existing works, as shown in Table \ref{tab:table1}.
We also compare our proposed method with closely-related baselines such as WAV2CLIP \cite{wu2022wav2clip}, AudioCLIP \cite{guzhov2022audioclip}, and ACL-SSL \cite{Park_2024_WACV}.

\subsubsection{Comparison on Standard Benchmarks}
In this section, a comparative analysis is conducted of our sound source localization method against existing approaches and established baselines.
Our evaluations follow a similar framework to previous methods \cite{chen2021localizing,mo2022localizing,senocak2022learning,sun2023learning}.
The model is trained on the VGGSound-144K dataset, with performance assessments carried out on the VGG-SS and SoundNet-Flickr test sets.
It is important to note that all compared models are trained with equivalent data volumes.
Our model, distinctively, does not utilize object-guided refinement (OGL).
The results are detailed in Table \ref{tab:table1}.

\begin{table}[!ht]
    \centering
    \begin{tabular}{lcccc}
        \toprule
        \multirow{2}{*}{\textbf{Method}} & \multicolumn{2}{c}{\textbf{VGG-SS}} & \multicolumn{2}{c}{\textbf{SoundNet-Flickr}} \\
         & $\mathbf{cIoU} \uparrow$ & $\mathbf{AUC} \uparrow$ & $\mathbf{cIoU} \uparrow$ & $\mathbf{AUC} \uparrow$ \\
        \midrule
        Attention & 18.50 & 30.20 & 66.00 & 55.80 \\
        CoarseToFine & 29.10 & 34.80 & - & - \\
        LCBM & 32.20 & 36.60 & - & - \\
        LVS & 34.40 & 38.20 & 71.90 & 58.20 \\
        HardPos & 34.60 & 38.00 & 76.80 & 59.20 \\
        SSPL & 33.90 & 38.00 & 76.70 & 60.50 \\
        EZ-VSL\dag & 35.96 & 38.20 & 78.31 & 61.74 \\
        EZ-VSL\ddag & 38.85 & 39.54 & 83.94 & 63.60 \\
        SSL-TIE & 38.63 & 39.65 & 79.50 & 61.20 \\
        SLAVC\dag & 37.79 & 39.40 & 83.60 & - \\
        SLAVC\ddag & 39.80 & - & \textbf{86.00} & - \\
        MarginNCE\dag & 38.25 & 39.06 & 83.94 & 63.20 \\
        MarginNCE\ddag & 39.78 & 40.01 & 85.14 & 64.55 \\
        HearTheFlow & 39.40 & 40.00 & 84.80 & 64.00 \\
        FNAC\dag & 39.50 & 39.66 & 84.73 & 63.76 \\
        FNAC\ddag& 41.85 & 40.80 & 85.14 & 64.30 \\
        Alignment\dag & 39.94 & 40.02 & 79.60 & 63.44 \\
        Alignment\ddag & 42.64 & 41.48 & 82.40 & 64.60 \\
        \midrule
        \multicolumn{5}{l}{\textit{Baseline:}} \\
        WAV2CLIP & 37.71 & 39.93 & 26.00 & 29.60 \\
        AudioCLIP & 44.15 & 46.23 & 47.20 & 45.22 \\
        ACL-SSL\dag & 49.46 & 46.32 & 80.80 & 64.62 \\
        \midrule
        \rowcolor{LightGreen} \textbf{\textsc{SouPLe}\dag}  & \textbf{53.21} & \textbf{48.15} & 84.80 & \textbf{67.64} \\
        & \textcolor{blue}{$\mathbf{3.75\uparrow}$} & \textcolor{blue}{$\mathbf{1.83\uparrow}$} & \textcolor{blue}{$\mathbf{4.00\uparrow}$} & \textcolor{blue}{$\mathbf{3.02\uparrow}$}\\
        \rowcolor{LightGreen} \textbf{\textsc{SouPLe}\dag-50ep}  & 54.76 & 49.40 & 83.60 & 65.76 \\
        \bottomrule
    \end{tabular}
    \caption{Quantitative results on the VGG-SS and SoundNet-Flickr test sets. All models were trained on 144K samples from VGGSound. SLAVC does not report AUC scores. The last row indicates improvement over ACL-SSL. $\dag$ and $\ddag$ denote without and with OGL, respectively.}
    \label{tab:table1}
\end{table}

We first compare our method with existing sound source localization approaches on the VGG-SS benchmark, as shown in Table \ref{tab:table1}.
It is also noteworthy that our use of CLIP does not involve explicit text input.
Compared with ACL-SSL, which directly uses the conventional prompt "a photo of a" with the audio-embedded token $[V_A]$, our prompt learning further enhances the performance and generalization.
The results confirm that instance-conditional prompts are more domain-generalizable.

\subsubsection{Open-Set Audio-Visual Localization}
\citet{chen2021localizing} introduced an open-set evaluation framework to gauge the generalization capabilities of sound source localization techniques.
This framework tests models on both categories included in the training data (Heard) and those not included (Unheard).
In this evaluation, 110 categories randomly chosen from the VGGSound dataset are designated for training, while a distinct set of 110 categories is reserved for testing, ensuring exposure to novel and unseen categories during the evaluation phase.
Experiments are conducted using the identical train/test split as established in prior studies \cite{chen2021localizing,mo2022localizing,park2023marginnce}.
Notably, our methodology diverges from previous ones as it does not employ object-guided refinement (OGL).
The results, detailed in Table \ref{tab:table2}, demonstrate that our approach significantly surpasses existing methods in localizing both Heard and Unheard categories.

\begin{table}[!htb]
    \centering
    \begin{tabular}{llcc}
        \toprule
        \textbf{Test Class} & \textbf{Method} & \textbf{cIoU} $\uparrow$ & \textbf{AUC} $\uparrow$ \\
        \midrule
        \multirow{11}{*}{ Heard 110} & LVS  & 28.90 & 36.20 \\
        & EZ-VSL\dag & 31.86 & 36.19 \\
        & EZ-VSL\ddag & 37.25 & 38.97 \\
        & SLAVC\dag & 35.84 & - \\
        & SLAVC\ddag & 38.22 & \\
        & FNAC\ddag & 39.54 & 39.83 \\
        & Alignment\dag  & 38.31 & 39.05 \\
        & Alignment (w OGL) & 41.85 & 40.93 \\
        & ACL-SSL\dag & 48.44 & 45.06 \\
        \midrule
        \rowcolor{LightGreen} & \textbf{\textsc{SouPLe}\dag} & \textbf{54.76} & \textbf{48.86} \\
        & & \textcolor{blue}{$\mathbf{6.32\uparrow}$} & \textcolor{blue}{$\mathbf{3.80\uparrow}$}\\
        \rowcolor{LightGreen} & \textbf{\textsc{SouPLe}\dag-50ep} & 54.85 & 48.98 \\
        \midrule
        \multirow{11}{*}{ Unheard 110} & LVS & 26.30 & 34.70 \\
        & EZ-VSL\dag  & 32.66 & 36.72 \\
        & EZ-VSL\ddag  & 39.57 & 39.60 \\
        & SLAVC\dag & 36.50 & - \\
        & SLAVC\ddag  & 38.87 & - \\
        & FNAC\ddag & 42.91 & 41.17 \\
        & Alignment\dag & 39.11 & 39.80 \\
        & Alignment (w OGL) & 42.94 & 41.54 \\
        & ACL-SSL\dag & 41.98 & 41.55 \\
        \midrule
        \rowcolor{LightGreen} & \textbf{\textsc{SouPLe}\dag} & \textbf{48.40} & \textbf{46.24} \\
        & & \textcolor{blue}{$\mathbf{6.42\uparrow}$} & \textcolor{blue}{$\mathbf{4.69\uparrow}$}\\
        \rowcolor{LightGreen} & \textbf{\textsc{SouPLe}\dag-50ep} & 48.40 & 46.24 \\
        \bottomrule
    \end{tabular}
    \caption{Open-set audio-visual localization results on the train/test splits of \cite{chen2021localizing,mo2022localizing,park2023marginnce}. The last row indicates improvements over ACL-SSL. $\dag$ and $\ddag$ denote without and with OGL, respectively. "50ep" indicates training for 50 epochs.}
    \label{tab:table2}
\end{table}

\subsubsection{AVSBench}
Additional experiments were conducted using the AVSBench S4 and MS3 datasets \cite{zhou2022audio} to demonstrate our model's precise sound localization capabilities.
These datasets are specifically designed for audio-visual correspondence identification at the pixel level, which is audio-visual segmentation.
In these experiments, models were trained on the VGGSound-144K dataset and subsequently tested on the AVSBench datasets in a zero-shot fashion.
The results, which are detailed in Table \ref{tab:table4}, reveal two significant insights.
Firstly, in single-object scenarios within the S4 dataset, our approach markedly surpasses the previous method by a substantial margin, utilizing solely audio-visual data without text supervision.
This underscores the efficacy of the prompt learning method in improving generalization capabilities for sound source localization and segmentation tasks.
Secondly, in multi-object scenarios within the MS3 dataset, our method, \textsc{SouPLe}, falls behind ACL-SSL.
The underperformance is attributed to the absence of GT label supervision, leading \textsc{SouPLe} to segment all potential objects within the frames, resulting in inaccurate segmentation masks.
It is also pertinent to acknowledge that the task of sound source localization is, in theory, an unlabeled challenge.

\begin{table}[!htb]
    \centering
    \begin{tabular}{lcccc}
        \toprule
        \textbf{Method} & \multicolumn{2}{c}{\textbf{S4}} & \multicolumn{2}{c}{\textbf{MS3}} \\
         \textbf{(w/o OGL)} & \textbf{mIoU} $\uparrow$ & \textbf{F-Score} $\uparrow$ & \textbf{mIoU} $\uparrow$ & \textbf{F-Score} $\uparrow$ \\
        \midrule
        SLAVC & 28.10 & 34.60 & 24.37 & 25.56 \\
        MarginNCE& 33.27 & 45.33 & 27.31 & 31.56 \\
        FNAC & 27.15 & 31.40 & 21.98 & 22.50 \\
        Alignment & 29.60 & 35.90 & - & - \\
        \midrule
        \textit{Baselines:} & & & & \\
        WAV2CLIP& 28.70 & 35.35 & 25.09 & 23.84 \\
        AudioCLIP & 36.57 & 42.15 & 27.06 & 26.48 \\
        ACL-SSL & 59.76 & 69.03 & \textbf{41.08} & \textbf{46.67} \\
        \midrule
        \rowcolor{LightGreen} \textbf{\textsc{SouPLe}} & \textbf{62.89} & \textbf{71.47} & 38.96 & 43.30 \\
        & \textcolor{blue}{$\mathbf{3.13\uparrow}$} & \textcolor{blue}{$\mathbf{2.44\uparrow}$} & \textcolor{red}{$\mathbf{2.12\downarrow}$} & \textcolor{red}{$\mathbf{3.37\downarrow}$}\\
        \rowcolor{LightGreen} \textbf{\textsc{SouPLe}-50ep} & 64.58 & 74.00 & 40.19 & 46.00 \\
        \bottomrule
    \end{tabular}
    \caption{Quantitative results on the AVSBench test sets. The last row indicates the improvement over ACL-SSL.}
    \label{tab:table4}
\end{table}

\begin{table*}[!ht]
    \centering
    \begin{tabular}{lcccccc}
        \toprule
        \multirow{2}{*}{\textbf{Method}} & \multicolumn{3}{c}{\textbf{Extended VGG-SS}} & \multicolumn{3}{c}{\textbf{Extended Flickr}} \\
        & \textbf{AP} $\uparrow$ & \textbf{max-F1} $\uparrow$ & \textbf{LocAcc} $\uparrow$ & \textbf{AP} $\uparrow$ & \textbf{max-F1} $\uparrow$ & \textbf{LocAcc} $\uparrow$ \\
        \midrule
        SLAVC\dag & 32.95 & 40.00 & 37.79 & 51.63 & 59.10 & 83.60 \\
        MarginNCE\dag & 30.58 & 36.80 & 38.25 & 57.99 & 61.80 & 83.94 \\
        FNAC\dag & 23.48 & 33.70 & 39.50 & 50.40 & 62.30 & 84.73 \\
        Alignment\dag & 34.73 & 40.70 & 39.94 & 64.43 & 66.90 & 79.60 \\
        WAV2CLIP & 26.67 & 33.00 & 37.71 & 20.99 & 24.80 & 29.60 \\
        AudioCLIP & 23.79 & 32.80 & 44.15 & 34.00 & 38.80 & 45.22 \\
        ACL-SSL\dag & 40.79 & 49.10 & 49.46 & 76.07 & 73.20 & 80.80 \\
        \midrule
        \rowcolor{LightGreen} \textbf{\textsc{SouPLe}\dag} & \textbf{44.77} & \textbf{53.00} & \textbf{53.21} & \textbf{80.25} & \textbf{83.11} & \textbf{84.80} \\
        & \textcolor{blue}{$\mathbf{3.98\uparrow}$} & \textcolor{blue}{$\mathbf{4.10\uparrow}$} & \textcolor{blue}{$\mathbf{3.66\uparrow}$} & \textcolor{blue}{$\mathbf{4.18\uparrow}$} & \textcolor{blue}{$\mathbf{7.91\uparrow}$} & \textcolor{blue}{$\mathbf{4.00\uparrow}$}\\
        \rowcolor{LightGreen} \textbf{\textsc{SouPLe}\dag-50ep} & 45.81 & 54.22 & 54.76 & 79.89 & 82.96 & 83.60 \\
        \bottomrule
    \end{tabular}
    \caption{Quantitative results on the Extended VGG-SS and Extended SoundNet-Flickr benchmarks. All models were trained on $144K$ samples from VGGSound. Results for prior approaches are from \citet{mo2022closer}. The last row indicates the improvement over ACL-SSL. $\dag$ and $\ddag$ denote without and with OGL, respectively.}
    \label{tab:table3}
\end{table*}

\subsubsection{Extended SoundNet-Flickr/VGG-SS}
Existing benchmarks typically consist of sounding objects/regions in the scene.
However, in reality, silent objects or off-screen audio are also common occurrences.
\citet{mo2022closer} propose a new evaluation that extends the existing benchmarks to include non-audible frames, non-visible sound sources, and mismatched audio-visual pairs.
In this evaluation scenario, it is expected that sound localization methods should not highlight an object/region if the audio and visual signals are mismatched.
The experiments were conducted using the extended SoundNet-Flickr/VGG-SS datasets in Table \ref{tab:table3}.
Surprisingly, \textsc{SouPLe} also surpasses previous methods significantly by a large margin.
These results indicate that our method provides a robust audio-visual semantic relationship that is suitable for a task such as a non-audible/non-visible sound source.

\section{Ablation Study}
\subsection{Context Length}
An ablation study on context length was conducted, examining 4, 8, and 16 context tokens.
To ensure a fair comparison, random initialization was employed for all context tokens.
The results, summarized in Table \ref{tab:ctx-length}, show that on the VGG-SS test set, \textsc{SouPLe} performs best with just four context tokens, suggesting that a higher number of context tokens may detrimentally affect performance, proving that increasing the parameter size is not the key.
Based on these results about the context length, we carry out the rest of the ablation study with only four context tokens on the VGG-SS dataset.

\begin{table}[!ht]
    \centering
    \begin{tabular}{l|cc}
    \toprule
        \textbf{\# ctx} & $\mathbf{cIoU} \uparrow$ & $\mathbf{AUC} \uparrow$ \\
        \midrule
        \rowcolor{LightGreen} \textbf{ctx=4} & \textbf{53.21} & \textbf{48.15}\\
        ctx=8 & 52.01 & 47.32\\
        ctx=16 & 51.08 & 46.93\\
    \bottomrule
    \end{tabular}
    \caption{Ablation study on context length on the VGG-SS dataset.}
    \label{tab:ctx-length}
\end{table}

\subsection{Audio-Embedded Token Position in the Prompt}
This section examines the impact of the position of the $[V_A]$ token on performance.
The $[V_A]$ token is randomly positioned within various context token prompts.
As indicated in Table \ref{tab:VA-position}, positioning $[V_A]$ at the start of the prompt notably diminishes performance, whereas situating it at the end enhances performance significantly.
\begin{table}[!ht]
    \centering
    \resizebox{\linewidth}{!}{
    \begin{tabular}{ll|ccc}
    \toprule
        \textbf{\# ctx} & \textbf{$\mathbf{V_A}$ index} & \textbf{Token Order} & $\mathbf{cIoU} \uparrow$ & $\mathbf{AUC} \uparrow$ \\
        \midrule
        ctx=4 & pos=1 & $\mathbf{[V_A]}[V_1][V_2][V_3][V_4]$ & 49.91 & 46.21\\
        ctx=4 & pos=2 & $[V_1]\mathbf{[V_A]}[V_2][V_3][V_4]$ & 50.71 & 46.42\\
        ctx=4 & pos=3 & $[V_1][V_2]\mathbf{[V_A]}[V_3][V_4]$ & 51.08 & 46.93\\
        ctx=4 & pos=4 & $[V_1][V_2][V_3]\mathbf{[V_A]}[V_4]$ & 52.41 & 47.72\\
        \rowcolor{LightGreen} \textbf{ctx=4} & \textbf{pos=5} & $[V_1][V_2][V_3][V_4]\mathbf{[V_A]}$ & \textbf{53.21} & \textbf{48.15}\\
    \bottomrule
    \end{tabular}
    }
    \caption{Ablation study on the position of $[V_A]$.}
    \label{tab:VA-position}
\end{table}

We observe that positioning $[V_A]$ at the end of the learnable context (pos=5) yields the highest performance, whereas placing it at the start (pos=1) notably diminishes results. This phenomenon can be explained by the CLIP text transformer’s causal (left-to-right) attention. By situating $[V_A]$ after the learnable context tokens $[V_1]\dots[V_M]$, the transformer layers can process the instance-conditional visual cues first, effectively priming the semantic space before reaching the audio-driven query. As a result, $[V_A]$ can aggregate richer image-conditioned context, leading to stronger audio-visual alignment.

\subsection{Longer Training}
To understand the impact of training duration, we conducted an ablation study by comparing different numbers of training epochs.
The results are recorded in Table \ref{tab:longer-training}.
Longer training further enhances the performance of \textsc{SouPLe}, resulting in a $cIoU$ of 54.76 and an $AUC$ of 49.40 for 50 epochs.
However, to keep the comparison fair with ACL-SSL, we based our analysis solely on the results of 20 epochs.

\begin{table}[!ht]
    \centering
    \begin{tabular}{lc|cc}
    \toprule
        \textbf{\# ctx} & \textbf{\# epochs} & $\mathbf{cIoU} \uparrow$ & $\mathbf{AUC} \uparrow$ \\
        \midrule
        ctx=4 & 20 & 53.21 & 48.15\\
        ctx=4 & 40 & 53.66 & 48.49\\
        \rowcolor{LightGreen} \textbf{ctx=4} & \textbf{50} & \textbf{54.76} & \textbf{49.40}\\
    \bottomrule
    \end{tabular}
    \caption{Ablation study on training duration.}
    \label{tab:longer-training}
\end{table}

\begin{figure*}[!ht]
    \includegraphics[width=\linewidth]{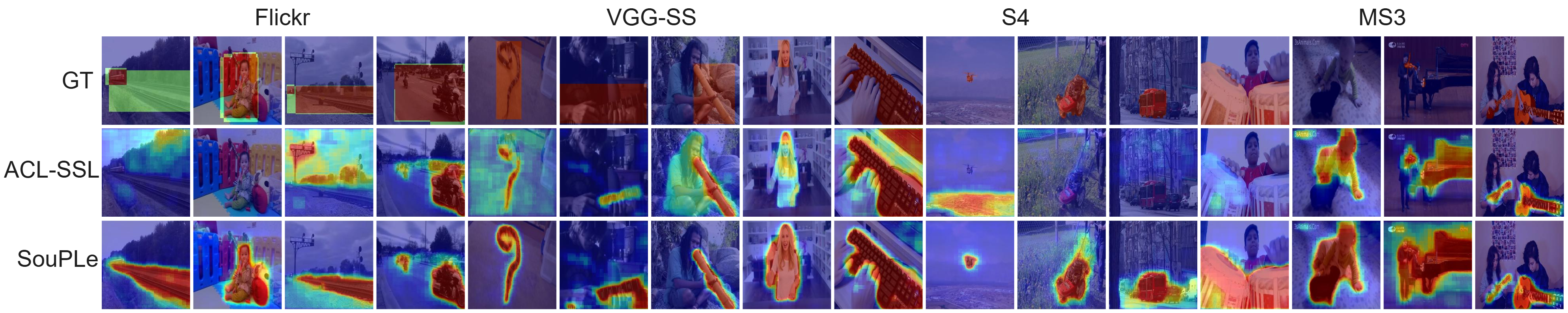}
    \caption{Sound localization results on VGG-SS, SoundNet-Flickr, and AVSBench, along with a comparison with ACL-SSL.}
    \label{fig:qualitative}
\end{figure*}

\subsection{Audio-Visual Feature Fusion}
Since previous methods have shown improvements when fusing the audio with visual features, in this section we also study its effect in \textsc{SouPLe}, the results are in Table \ref{tab:fusion}.

\begin{table}[!ht]
    \centering
    \begin{tabular}{lcc|cc}
    \toprule
        \textbf{Method} & \textbf{Fusion?} & \textbf{Ensemble?} & $\mathbf{cIoU} \uparrow$ & $\mathbf{AUC} \uparrow$ \\
        \midrule
        \rowcolor{LightGreen}\multirow{3}{*}{\textsc{SouPLe}} &  &  & \textbf{53.21} & \textbf{48.15}\\
         & \checkmark &  & 49.93 & 45.98\\
         &  & \checkmark  & 33.63 & 34.49\\
    \bottomrule
    \end{tabular}
    \caption{Ablation study on audio-visual feature fusion.}
    \label{tab:fusion}
\end{table}
We carried out three distinct experiments: one with the standard \textsc{SouPLe}, another with \textsc{SouPLe} incorporating fused audio and visual features, and a third with \textsc{SouPLe} employing an ensemble strategy.
In the fusion scenario, audio and visual features were combined prior to their introduction to the Meta-net, resulting in the final prompt being $\{V_1{\left(F_I+F_A\right)}, V_2{\left(F_I+F_A\right)},\dots, V_M{\left(F_I+F_A\right)}\}$.
Consequently, $[V_A]$ was omitted since the audio features had already been integrated with the visual features.
For the ensemble scenario, we altered the position of $[V_A]$ sequentially, as delineated in Table \ref{tab:VA-position}.
Instead of relying on a single position, we computed the mean across five different positions.
The outcomes revealed that while the fusion of audio-visual features can enhance the efficacy of other methods, it does not yield the same success when applied to \textsc{SouPLe}.
Furthermore, the performance was significantly diminished when multiple prompts were ensemble with varying $[V_A]$ positions.

\subsection{Failure Analysis on the MS3 Dataset}
\label{sec:failures-ms3}
\begin{figure}[!ht]
    \includegraphics[width=\linewidth]{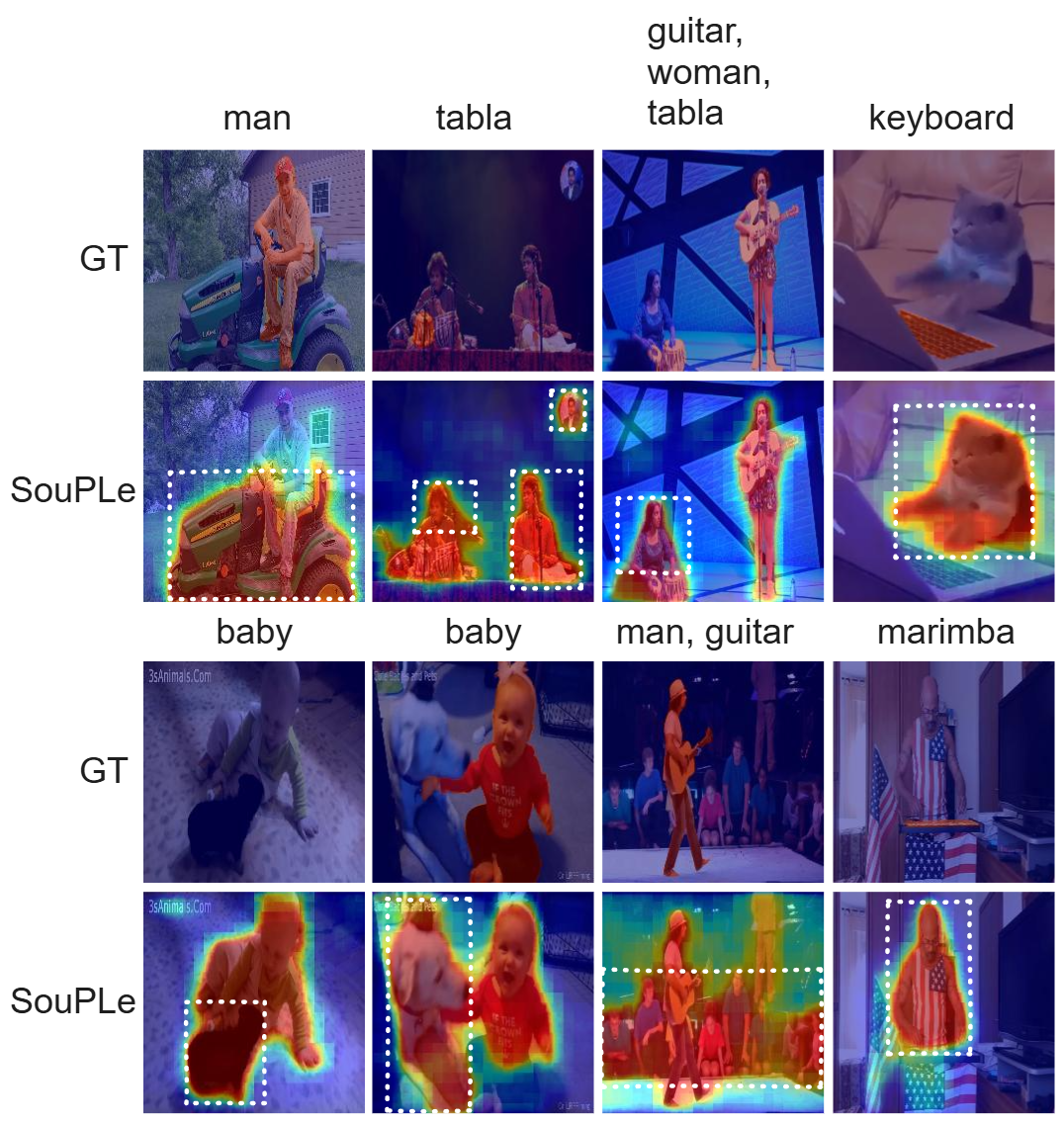}
    \caption{Failure cases visualization of \textsc{SouPLe} on MS3.}
    \label{fig:failures}
\end{figure}
In multi-object scenarios within the MS3 dataset, \textsc{SouPLe} falls behind ACL-SSL. We attribute this performance gap to the absence of class-specific supervision in our textless, label-free setting: when multiple plausible sounding objects co-occur, the model may segment several semantically relevant regions, which can be penalized when the annotations do not fully capture all sounding sources. Nevertheless, the results still indicate that prompt learning improves generalization in single-source settings and remains competitive in more challenging multi-source scenes.

Figure \ref{fig:failures} illustrates several representative failure cases from the MS3 dataset. In these examples, \textsc{SouPLe} sometimes highlights multiple candidate regions that are semantically consistent with the audio, whereas the ground-truth annotations correspond to only one of the objects. This phenomenon is particularly evident in complex scenes containing multiple interacting objects or overlapping sound sources. Despite these challenges, the results indicate that prompt learning improves generalization in single-source scenarios and remains competitive in more challenging multi-source environments.

\subsection{Qualitative Results}
Figure \ref{fig:qualitative} illustrates the comparative analysis between our method and ACL-SSL.
The visual examples show that our approach provides more accurate and finely localized regions for sounding objects than ACL-SSL, which often produces vague or misclassified segmentation areas.
In MS3, \textsc{SouPLe} often highlights multiple semantically plausible sounding objects, some of which may not be fully captured by the available annotations.
In summary, \textsc{SouPLe} demonstrates consistency in handling sound source objects of various sizes, regardless of their location and resolution.
These results are consistent with the quantitative data presented in Table \ref{tab:table1} and Table \ref{tab:table3}, further validating the superiority of our method over ACL-SSL and other prior techniques across all tested datasets.
\section{Conclusion}
\label{sec:conclusion}
We presented a prompt learning approach for sound source localization and segmentation. Our goal was to improve generalization on unlabeled datasets and previously unseen objects by replacing a fixed handcrafted prompt with an instance-conditional form. To this end, we introduced \textsc{SouPLe}, which generates learnable context tokens from visual features and combines them with an audio-embedded token to better preserve the semantic relationship between audio and visual inputs. Experimental results demonstrate that this simple prompt reformulation consistently improves audio-visual localization and segmentation, highlighting the potential of prompt learning for broader audio-visual understanding tasks.

\section*{Acknowledgements}
This work was supported by the National Cancer Center Grant(NCC-23113503, NCC-25104802).
{
    \small
    \bibliographystyle{ieeenat_fullname}
    \bibliography{main}

@String(CVPR= {IEEE Conf. Comput. Vis. Pattern Recog.})

@String(ICCV= {Int. Conf. Comput. Vis.})

@String(ECCV= {Eur. Conf. Comput. Vis.})

@String(ICASSP=	{ICASSP})

@String(CVPR  = {CVPR})

@String(ICCV  = {ICCV})

@String(ECCV  = {ECCV})

@inproceedings{chen2021localizing,
  title={Localizing visual sounds the hard way},
  author={Chen, Honglie and Xie, Weidi and Afouras, Triantafyllos and Nagrani, Arsha and Vedaldi, Andrea and Zisserman, Andrew},
  booktitle={Proceedings of the IEEE/CVF Conference on Computer Vision and Pattern Recognition},
  pages={16867--16876},
  year={2021}
}

@inproceedings{chen2020vggsound,
  title={Vggsound: A large-scale audio-visual dataset},
  author={Chen, Honglie and Xie, Weidi and Vedaldi, Andrea and Zisserman, Andrew},
  booktitle={ICASSP 2020-2020 IEEE International Conference on Acoustics, Speech and Signal Processing (ICASSP)},
  pages={721--725},
  year={2020},
  organization={IEEE}
}

@article{hu2020discriminative,
  title={Discriminative sounding objects localization via self-supervised audiovisual matching},
  author={Hu, Di and Qian, Rui and Jiang, Minyue and Tan, Xiao and Wen, Shilei and Ding, Errui and Lin, Weiyao and Dou, Dejing},
  journal={Advances in Neural Information Processing Systems},
  volume={33},
  pages={10077--10087},
  year={2020}
}

@article{lin2023unsupervised,
  title={Unsupervised sound localization via iterative contrastive learning},
  author={Lin, Yan-Bo and Tseng, Hung-Yu and Lee, Hsin-Ying and Lin, Yen-Yu and Yang, Ming-Hsuan},
  journal={Computer Vision and Image Understanding},
  volume={227},
  pages={103602},
  year={2023},
  publisher={Elsevier}
}

@inproceedings{liu2022exploiting,
  title={Exploiting transformation invariance and equivariance for self-supervised sound localisation},
  author={Liu, Jinxiang and Ju, Chen and Xie, Weidi and Zhang, Ya},
  booktitle={Proceedings of the 30th ACM International Conference on Multimedia},
  pages={3742--3753},
  year={2022}
}

@inproceedings{qian2020multiple,
  title={Multiple sound sources localization from coarse to fine},
  author={Qian, Rui and Hu, Di and Dinkel, Heinrich and Wu, Mengyue and Xu, Ning and Lin, Weiyao},
  booktitle={Computer Vision--ECCV 2020: 16th European Conference, Glasgow, UK, August 23--28, 2020, Proceedings, Part XX 16},
  pages={292--308},
  year={2020},
  organization={Springer}
}

@inproceedings{radford2021learning,
  title={Learning transferable visual models from natural language supervision},
  author={Radford, Alec and Kim, Jong Wook and Hallacy, Chris and Ramesh, Aditya and Goh, Gabriel and Agarwal, Sandhini and Sastry, Girish and Askell, Amanda and Mishkin, Pamela and Clark, Jack and others},
  booktitle={International conference on machine learning},
  pages={8748--8763},
  year={2021},
  organization={PMLR}
}

@inproceedings{senocak2018learning,
  title={Learning to localize sound source in visual scenes},
  author={Senocak, Arda and Oh, Tae-Hyun and Kim, Junsik and Yang, Ming-Hsuan and Kweon, In So},
  booktitle={Proceedings of the IEEE Conference on Computer Vision and Pattern Recognition},
  pages={4358--4366},
  year={2018}
}

@inproceedings{song2022self,
  title={Self-supervised predictive learning: A negative-free method for sound source localization in visual scenes},
  author={Song, Zengjie and Wang, Yuxi and Fan, Junsong and Tan, Tieniu and Zhang, Zhaoxiang},
  booktitle={Proceedings of the IEEE/CVF Conference on Computer Vision and Pattern Recognition},
  pages={3222--3231},
  year={2022}
}

@inproceedings{zhou2022audio,
  title={Audio--visual segmentation},
  author={Zhou, Jinxing and Wang, Jianyuan and Zhang, Jiayi and Sun, Weixuan and Zhang, Jing and Birchfield, Stan and Guo, Dan and Kong, Lingpeng and Wang, Meng and Zhong, Yiran},
  booktitle={European Conference on Computer Vision},
  pages={386--403},
  year={2022},
  organization={Springer}
}

@article{wang2023prompting,
  title={Prompting Segmentation with Sound is Generalizable Audio-Visual Source Localizer},
  author={Wang, Yaoting and Liu, Weisong and Li, Guangyao and Ding, Jian and Hu, Di and Li, Xi},
  journal={arXiv preprint arXiv:2309.07929},
  year={2023}
}

@inproceedings{zhou2022conditional,
  title={Conditional prompt learning for vision-language models},
  author={Zhou, Kaiyang and Yang, Jingkang and Loy, Chen Change and Liu, Ziwei},
  booktitle={Proceedings of the IEEE/CVF conference on computer vision and pattern recognition},
  pages={16816--16825},
  year={2022}
}

@inproceedings{arandjelovic2018objects,
  title={Objects that sound},
  author={Arandjelovic, Relja and Zisserman, Andrew},
  booktitle={Proceedings of the European conference on computer vision (ECCV)},
  pages={435--451},
  year={2018}
}

@inproceedings{bhati2023segmental,
  title={Segmental speechclip: Utilizing pretrained image-text models for audio-visual learning},
  author={Bhati, Saurabhchand and Villalba, Jes{\'u}s and Moro-Velazquez, Laureano and Thebaud, Thomas and Dehak, Najim},
  booktitle={INTERSPEECH},
  year={2023}
}

@article{chen2022beats,
  title={Beats: Audio pre-training with acoustic tokenizers},
  author={Chen, Sanyuan and Wu, Yu and Wang, Chengyi and Liu, Shujie and Tompkins, Daniel and Chen, Zhuo and Wei, Furu},
  journal={arXiv preprint arXiv:2212.09058},
  year={2022}
}

@inproceedings{chen2021exploring,
  title={Exploring simple siamese representation learning},
  author={Chen, Xinlei and He, Kaiming},
  booktitle={Proceedings of the IEEE/CVF conference on computer vision and pattern recognition},
  pages={15750--15758},
  year={2021}
}

@article{dong2022clipsep,
  title={Clipsep: Learning text-queried sound separation with noisy unlabeled videos},
  author={Dong, Hao-Wen and Takahashi, Naoya and Mitsufuji, Yuki and McAuley, Julian and Berg-Kirkpatrick, Taylor},
  journal={arXiv preprint arXiv:2212.07065},
  year={2022}
}

@article{fan2024revisit,
  title={Revisit weakly-supervised audio-visual video parsing from the language perspective},
  author={Fan, Yingying and Wu, Yu and Du, Bo and Lin, Yutian},
  journal={Advances in Neural Information Processing Systems},
  volume={36},
  year={2024}
}

@inproceedings{guzhov2022audioclip,
  title={Audioclip: Extending clip to image, text and audio},
  author={Guzhov, Andrey and Raue, Federico and Hees, J{\"o}rn and Dengel, Andreas},
  booktitle={ICASSP 2022-2022 IEEE International Conference on Acoustics, Speech and Signal Processing (ICASSP)},
  pages={976--980},
  year={2022},
  organization={IEEE}
}

@inproceedings{jia2021scaling,
  title={Scaling up visual and vision-language representation learning with noisy text supervision},
  author={Jia, Chao and Yang, Yinfei and Xia, Ye and Chen, Yi-Ting and Parekh, Zarana and Pham, Hieu and Le, Quoc and Sung, Yun-Hsuan and Li, Zhen and Duerig, Tom},
  booktitle={International conference on machine learning},
  pages={4904--4916},
  year={2021},
  organization={PMLR}
}

@article{li2021space,
  title={Space-time memory network for sounding object localization in videos},
  author={Li, Sizhe and Tian, Yapeng and Xu, Chenliang},
  journal={arXiv preprint arXiv:2111.05526},
  year={2021}
}

@inproceedings{luddecke2022image,
  title={Image segmentation using text and image prompts},
  author={L{\"u}ddecke, Timo and Ecker, Alexander},
  booktitle={Proceedings of the IEEE/CVF conference on computer vision and pattern recognition},
  pages={7086--7096},
  year={2022}
}

@inproceedings{mahmud2023ave,
  title={Ave-clip: Audioclip-based multi-window temporal transformer for audio visual event localization},
  author={Mahmud, Tanvir and Marculescu, Diana},
  booktitle={Proceedings of the IEEE/CVF Winter Conference on Applications of Computer Vision},
  pages={5158--5167},
  year={2023}
}

@article{mo2022closer,
  title={A closer look at weakly-supervised audio-visual source localization},
  author={Mo, Shentong and Morgado, Pedro},
  journal={Advances in Neural Information Processing Systems},
  volume={35},
  pages={37524--37536},
  year={2022}
}

@inproceedings{mo2022localizing,
  title={Localizing visual sounds the easy way},
  author={Mo, Shentong and Morgado, Pedro},
  booktitle={European Conference on Computer Vision},
  pages={218--234},
  year={2022},
  organization={Springer}
}

@inproceedings{oya2020we,
  title={Do we need sound for sound source localization?},
  author={Oya, Takashi and Iwase, Shohei and Natsume, Ryota and Itazuri, Takahiro and Yamaguchi, Shugo and Morishima, Shigeo},
  booktitle={Proceedings of the Asian Conference on Computer Vision},
  year={2020}
}

@inproceedings{park2023marginnce,
  title={MarginNCE: Robust sound localization with a negative margin},
  author={Park, Sooyoung and Senocak, Arda and Chung, Joon Son},
  booktitle={ICASSP 2023-2023 IEEE International Conference on Acoustics, Speech and Signal Processing (ICASSP)},
  pages={1--5},
  year={2023},
  organization={IEEE}
}

@inproceedings{senocak2022learning,
  title={Learning sound localization better from semantically similar samples},
  author={Senocak, Arda and Ryu, Hyeonggon and Kim, Junsik and Kweon, In So},
  booktitle={ICASSP 2022-2022 IEEE International Conference on Acoustics, Speech and Signal Processing (ICASSP)},
  pages={4863--4867},
  year={2022},
  organization={IEEE}
}

@inproceedings{senocak2022less,
  title={Less can be more: Sound source localization with a classification model},
  author={Senocak, Arda and Ryu, Hyeonggon and Kim, Junsik and Kweon, In So},
  booktitle={Proceedings of the IEEE/CVF Winter Conference on Applications of Computer Vision},
  pages={3308--3317},
  year={2022}
}

@inproceedings{senocak2023sound,
  title={Sound source localization is all about cross-modal alignment},
  author={Senocak, Arda and Ryu, Hyeonggon and Kim, Junsik and Oh, Tae-Hyun and Pfister, Hanspeter and Chung, Joon Son},
  booktitle={Proceedings of the IEEE/CVF International Conference on Computer Vision},
  pages={7777--7787},
  year={2023}
}

@inproceedings{sun2023learning,
  title={Learning audio-visual source localization via false negative aware contrastive learning},
  author={Sun, Weixuan and Zhang, Jiayi and Wang, Jianyuan and Liu, Zheyuan and Zhong, Yiran and Feng, Tianpeng and Guo, Yandong and Zhang, Yanhao and Barnes, Nick},
  booktitle={Proceedings of the IEEE/CVF Conference on Computer Vision and Pattern Recognition},
  pages={6420--6429},
  year={2023}
}

@inproceedings{tan2023language,
  title={Language-guided audio-visual source separation via trimodal consistency},
  author={Tan, Reuben and Ray, Arijit and Burns, Andrea and Plummer, Bryan A and Salamon, Justin and Nieto, Oriol and Russell, Bryan and Saenko, Kate},
  booktitle={Proceedings of the IEEE/CVF Conference on Computer Vision and Pattern Recognition},
  pages={10575--10584},
  year={2023}
}

@inproceedings{tian2018audio,
  title={Audio-visual event localization in unconstrained videos},
  author={Tian, Yapeng and Shi, Jing and Li, Bochen and Duan, Zhiyao and Xu, Chenliang},
  booktitle={Proceedings of the European conference on computer vision (ECCV)},
  pages={247--263},
  year={2018}
}

@inproceedings{wu2022wav2clip,
  title={Wav2clip: Learning robust audio representations from clip},
  author={Wu, Ho-Hsiang and Seetharaman, Prem and Kumar, Kundan and Bello, Juan Pablo},
  booktitle={ICASSP 2022-2022 IEEE International Conference on Acoustics, Speech and Signal Processing (ICASSP)},
  pages={4563--4567},
  year={2022},
  organization={IEEE}
}

@inproceedings{xuan2022proposal,
  title={A proposal-based paradigm for self-supervised sound source localization in videos},
  author={Xuan, Hanyu and Wu, Zhiliang and Yang, Jian and Yan, Yan and Alameda-Pineda, Xavier},
  booktitle={Proceedings of the IEEE/CVF Conference on Computer Vision and Pattern Recognition},
  pages={1029--1038},
  year={2022}
}

@article{yariv2023audiotoken,
  title={Audiotoken: Adaptation of text-conditioned diffusion models for audio-to-image generation},
  author={Yariv, Guy and Gat, Itai and Wolf, Lior and Adi, Yossi and Schwartz, Idan},
  journal={arXiv preprint arXiv:2305.13050},
  year={2023}
}

@InProceedings{Park_2024_WACV,
    author    = {Park, Sooyoung and Senocak, Arda and Chung, Joon Son},
    title     = {Can CLIP Help Sound Source Localization?},
    booktitle = {Proceedings of the IEEE/CVF Winter Conference on Applications of Computer Vision (WACV)},
    month     = {January},
    year      = {2024},
    pages     = {5711-5720}
}

@article{zhou2022learning,
  title={Learning to prompt for vision-language models},
  author={Zhou, Kaiyang and Yang, Jingkang and Loy, Chen Change and Liu, Ziwei},
  journal={International Journal of Computer Vision},
  volume={130},
  number={9},
  pages={2337--2348},
  year={2022},
  publisher={Springer}
}

@inproceedings{khattakMaPLe,
    title={MaPLe: Multi-modal Prompt Learning},
    author={khattak, Muhammad Uzair and Rasheed, Hanoona and Maaz, Muhammad and Khan, Salman and Khan, Fahad Shahbaz},
    booktitle={The IEEE/CVF Conference on Computer Vision and Pattern Recognition},
    year={2023}
}

@InProceedings{Khattak_2023_ICCV,
    author    = {Khattak, Muhammad Uzair and Wasim, Syed Talal and Naseer, Muzammal and Khan, Salman and Yang, Ming-Hsuan and Khan, Fahad Shahbaz},
    title     = {Self-regulating Prompts: Foundational Model Adaptation without Forgetting},
    booktitle = {Proceedings of the IEEE/CVF International Conference on Computer Vision (ICCV)},
    month     = {October},
    year      = {2023},
    pages     = {15190-15200}
}

@InProceedings{lueddecke22_cvpr,
    author    = {L\"uddecke, Timo and Ecker, Alexander},
    title     = {Image Segmentation Using Text and Image Prompts},
    booktitle = {Proceedings of the IEEE/CVF Conference on Computer Vision and Pattern Recognition (CVPR)},
    month     = {June},
    year      = {2022},
    pages     = {7086-7096}
}

@misc{nguyen2024save,
    title={SAVE: Segment Audio-Visual Easy way using Segment Anything Model},
    author={Khanh-Binh Nguyen and Chae Jung Park},
    year={2024},
    eprint={2407.02004},
    archivePrefix={arXiv},
    primaryClass={cs.CV}
}

@ARTICLE{10870225,
  author={Nguyen, Khanh-Binh and Park, Chae Jung},
  journal={IEEE Access}, 
  title={On Calibration of Prompt Learning Using Temperature Scaling}, 
  year={2025},
  volume={},
  number={},
  pages={1-1},
  doi={10.1109/ACCESS.2025.3538617}}
}


\end{document}